\pdfoutput=1

\documentclass[11pt]{article}

\usepackage{ACL2023}
\usepackage{dsfont}
\usepackage{times}
\usepackage{latexsym}
\usepackage{svg}
\usepackage[T1]{fontenc}
\usepackage{caption}
\usepackage[utf8]{inputenc}
\usepackage{pgfplots}
\usepackage{microtype}

\usepackage{algorithm,algpseudocode}

\usepackage{makecell}
\usepackage{times}
\usepackage{latexsym}
\usepackage{amsmath}
\usepackage[T1]{fontenc}

\usepackage[utf8]{inputenc}

\usepackage{microtype}
\usepackage{graphicx}
\usepackage{booktabs}
\usepackage{inconsolata}
\usepackage{amsfonts}
\usepackage{pifont}
\usepackage{multirow}
\usepackage{supertabular}
\usepackage{tikz}
\usepackage{array}
\usepackage{colortbl}
\newcommand{\partialvrule}[1]{%
  \tikz[overlay]{\draw (0,-1.1ex) -- (0,2.5ex);}%
}
\title{MIKE: A New Benchmark for
Fine-grained Multimodal Entity Knowledge Editing}

\author{
  Jiaqi Li$^{1,3}$, 
  Miaozeng Du$^{1,3}$, 
  Chuanyi Zhang$^2$, 
  Yongrui Chen$^{1,3}$, 
  Nan Hu$^{1,3}$, \\
  \textbf{Guilin Qi}$^{1,3}$, 
  \textbf{Haiyun Jiang}$^4$, 
  \textbf{Siyuan Cheng}$^5$, 
  \textbf{Bozhong Tian}$^5$
}
\begin{document}
\maketitle
\begin{abstract}


Multimodal knowledge editing represents a critical advancement in enhancing the capabilities of Multimodal Large Language Models (MLLMs). Despite its potential, current benchmarks predominantly focus on coarse-grained knowledge, leaving the intricacies of fine-grained (FG) multimodal entity knowledge largely unexplored. This gap presents a notable challenge, as FG entity recognition is pivotal for the practical deployment and effectiveness of MLLMs in diverse real-world scenarios. To bridge this gap, we introduce MIKE, a comprehensive benchmark and dataset specifically designed for the FG \textbf{m}ultimodal ent\textbf{i}ty \textbf{k}nowledge \textbf{e}diting.  MIKE encompasses a suite of tasks tailored to assess different perspectives, including Vanilla Name Answering, Entity-Level Caption, and Complex-Scenario Recognition. In addition, a new form of knowledge editing, Multi-Step Editing, is introduced to evaluate the editing efficiency. Through our extensive evaluations, we demonstrate that the current state-of-the-art methods face significant challenges in tackling our proposed benchmark, underscoring the complexity of FG knowledge editing in MLLMs. Our findings spotlight the urgent need for novel approaches in this domain, setting a clear agenda for future research and development efforts within the community.


\end{abstract}

\section{Introduction}

Multimodal knowledge editing (MKE) \cite{DBLP:conf/emnlp/YaoWT0LDC023,DBLP:journals/corr/abs-2401-01286,DBLP:conf/nips/MengBAB22,DBLP:conf/emnlp/DongDSXSL22,DBLP:conf/eacl/HaseDCLKSBI23,DBLP:conf/iclr/MengSABB23} plays a critical role in maintaining and improving the accuracy of Multimodal Large Language Models (MLLMs) \cite{DBLP:journals/corr/abs-2304-08485,DBLP:journals/corr/abs-2305-03726,DBLP:conf/nips/AlayracDLMBHLMM22,DBLP:conf/icml/0008LSH23}. Central to MKE is the capability to update outdated, unknown, or incorrect knowledge within MLLMs. Recent developments in this field, such as the benchmark MMEdit proposed by \citet{DBLP:conf/emnlp/0008TL0WC023}, signify considerable progress. Drawing from datasets of Visual Question Answering (VQA) \cite{DBLP:journals/corr/abs-2211-09699,DBLP:conf/cvpr/0001KS00C23} and Image Caption \cite{DBLP:conf/icml/0008LSH23,DBLP:conf/cvpr/Ramos0EK23} tasks, MMEdit offers a platform to test the editability of MLLMs. However, a critical issue remains in its primary focus on coarse-grained knowledge, which often falls short of accurately representing real-world fine-grained (FG) entities and scenarios.
\begin{figure}[t]
\centering  
\includegraphics[width=\linewidth]{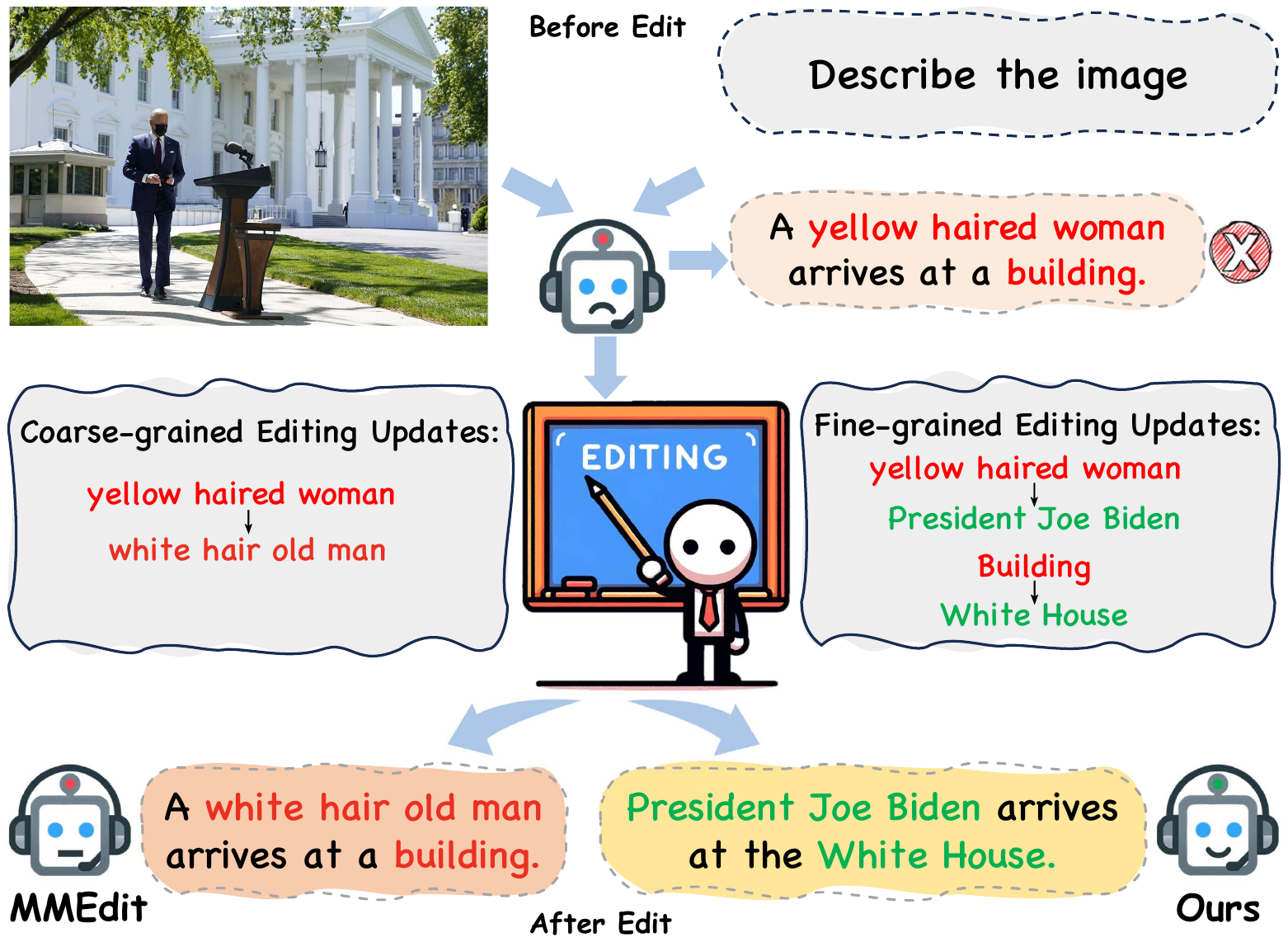}    
\vspace{-0.7cm}
\caption{The comparison between MMEdit \cite{DBLP:conf/emnlp/0008TL0WC023} and ours (MIKE). MIKE focuses on editing fine-grained multimodal entity knowledge.}
\label{figintro}
\end{figure}

To underscore the limitations of a coarse-grained focus, consider a real-life example in political image captioning as shown in Figure \ref{figintro}. An ideal MLLM output would be a fine-grained and specific caption like \emph{"President Joe Biden arrives at the White House".} However, a coarse-grained approach might yield a nondescript caption such as \emph{"A white hair old man arrives at a building".} This lack of specificity fails to capture the critical details and convey key information to the users of MLLMs, illustrating how FG entity recognition is essential for delivering accurate information.


While the necessity for more detailed, entity-specific information is clear, editing FG knowledge into MLLMs is a complex and challenging endeavor. Traditional FG image classification tasks \cite{DBLP:conf/cvpr/WeiFS0GY23, DBLP:conf/cvpr/TangYC23, DBLP:conf/cvpr/GuoLRGM023} demand vision encoders to discern and categorize visually similar items. The task becomes even more difficult when extending to MLLMs. MLLMs are required to not only recognize FG visual entities but also to understand and map them to corresponding textual descriptions. Although recent studies \cite{DBLP:conf/emnlp/ChenHLSCRC23, DBLP:journals/corr/abs-2302-11154} have demonstrated a nascent ability in MLLMs to identify multimodal knowledge at the entity level, their performance notably lags in handling FG entities as compared to coarse-grained ones. This performance gap highlights the substantial challenges in accurately recognizing FG entities by MLLMs.

Given these challenges, the question remains: Can we effectively edit FG multimodal entity knowledge into MLLMs?  Addressing this query is not only crucial for advancing the field of MLLMs but also for unlocking a myriad of applications requiring detailed understanding. To explore this problem, we propose a comprehensive and challenging benchmark for fine-grained \textbf{m}ultimodal \textbf{e}ntity \textbf{k}nowledge \textbf{e}diting (MIKE). It is composed of more than 1000 FG entities, each of which includes at least 5 images. To challenge MKE methods and meet the needs of real scenes, we purposefully create a diverse set of tasks from different angles: (i) Vanilla Name Answering, where MLLMs are required to answer the short name of the entity in the image; (ii) Entity-Level Caption, where MLLMs need to caption the image not only the general content but the entity name as well;  (iii) Complex-Scenario Recognition, where MLLMs need to recognize a targeted entity under a complex visual field of multiple entities. In addition, extending the normal knowledge editing form, we propose Multi-Step Editing. In this form, MLLMs are edited with 2-4 FG entity images instead of one. We utilize EasyEdit toolkit \cite{DBLP:journals/corr/abs-2308-07269} to assess several knowledge editing approaches on MIKE. For the evaluation, we propose entity-oriented metrics under the setting of Reliability, Generality and Locality. Through extensive experiments, we find (i) each editing method exhibits specific limitations; (ii) the most challenging task for current editing methods is Entity-level Caption; (iii) different generality tasks affect the ability of MKE in some aspects; (iv)
model size does not matter. For a detailed discussion of these findings and additional results, please refer to Section \ref{exp}.

We summarize main contributions as follows:

\textbullet\ A novel multimodal knowledge editing benchmark, called MIKE, is introduced. Compared with existing benchmark, MIKE focuses on editing fine-grained multimodal entities into MLLMs. To the best of our knowledge, we are the first to explore fine-grained multimodal entities in multimodal knowledge editing.

\textbullet\ To test multimodal knowledge editing methods, we design three challenging tasks: Vanilla Name Answering, Entity-level Caption and Complex-Scenario Recognition. These tasks could significantly meet real-world applications.

\textbullet\ We propose a Multi-Step Editing form for editing fine-grained multimodal entities. Extensive results show the improvement and effects of different number of editing images.

\section{Related Work}

\subsection{Knowledge Editing}

The world is changing all the time, but the training data of a particular model is fixed during training. If the model can not learn online, the knowledge inside the model will be outdated. As retraining is expensive most of the time, knowledge editing methods \cite{DBLP:conf/emnlp/YaoWT0LDC023,DBLP:journals/corr/abs-2401-01286} are needed to edit the model after training and modify the knowledge in it. One way to update the model's knowledge is through fine-tuning. However, to minimize the loss of previously learned knowledge, certain restrictions need to be imposed during fine-tuning. \citet{DBLP:journals/corr/abs-2012-00363} minimizes the loss of editing target knowledge when the loss of non-editing target knowledge is less than a minimal value $\delta$. \citet{DBLP:conf/nips/TannoPNL22} draw on the Bayesian view of knowledge editing. Another way is to store the new or corrected knowledge in the form of a patch model, alongside the original model, and utilize them together. Mend \cite{DBLP:conf/iclr/MitchellLBFM22} and KE \cite{DBLP:conf/emnlp/CaoAT21}  train
a hypernetwork to learn the gradient of edited parameters when encoding new knowledge. SERAC \cite{DBLP:conf/icml/MitchellLBMF22} trains a BERT \cite{DBLP:conf/naacl/DevlinCLT19} classifier as a scope classifier and a T5 \cite{DBLP:journals/jmlr/RaffelSRLNMZLL20} as a Counterfactual model based on the new knowledge data. In addition, a more explanatory idea is locate-then-edit\cite{DBLP:conf/nips/MengBAB22,DBLP:conf/emnlp/DongDSXSL22,DBLP:conf/eacl/HaseDCLKSBI23,DBLP:conf/iclr/MengSABB23}. According to different prompts that express the same meaning, they locate the neurons that store the corresponding knowledge and modify their value. Recently, \citet{DBLP:conf/emnlp/ZhengLDFWXC23} investigated the potential of using in-context learning in knowledge editing of LLMs. The proposed IKE method achieves a competitive knowledge editing effect without any parameter modification.

\subsection{Multimodal Large Language Models}

Typically, Multimodal Large Language Models are structured by combining a visual encoder with a language model, with the two components linked via a connector. \citet{DBLP:conf/nips/AlayracDLMBHLMM22} introduce a novel approach which utilizes a query-based cross-attention mechanism. This groundbreaking technique creates a resilient vision-language interactive module. BLIP-2 \cite{DBLP:conf/icml/0008LSH23} substitute the cross-attention with a Q-Former, which is a lightweight Transformer architecture. MiniGPT-4 \cite{DBLP:journals/corr/abs-2304-10592} and InstructBLIP both improve the BLIP-2 performance by incorporating instruction tuning datasets gathered from varied public datasets. LLaVA and Otter \cite{DBLP:journals/corr/abs-2304-08485,DBLP:journals/corr/abs-2305-03726} design a suit of instruction data system to enhance the understanding ability. Compared with previous training stages, \citet{DBLP:journals/corr/abs-2308-12966} propose a three-stage training process to further align the multimodal representations. CogVLM \cite{DBLP:journals/corr/abs-2311-03079} introduces a visual expert to boost the performance.

\begin{figure}[t]
\centering
\includegraphics[width=\linewidth]{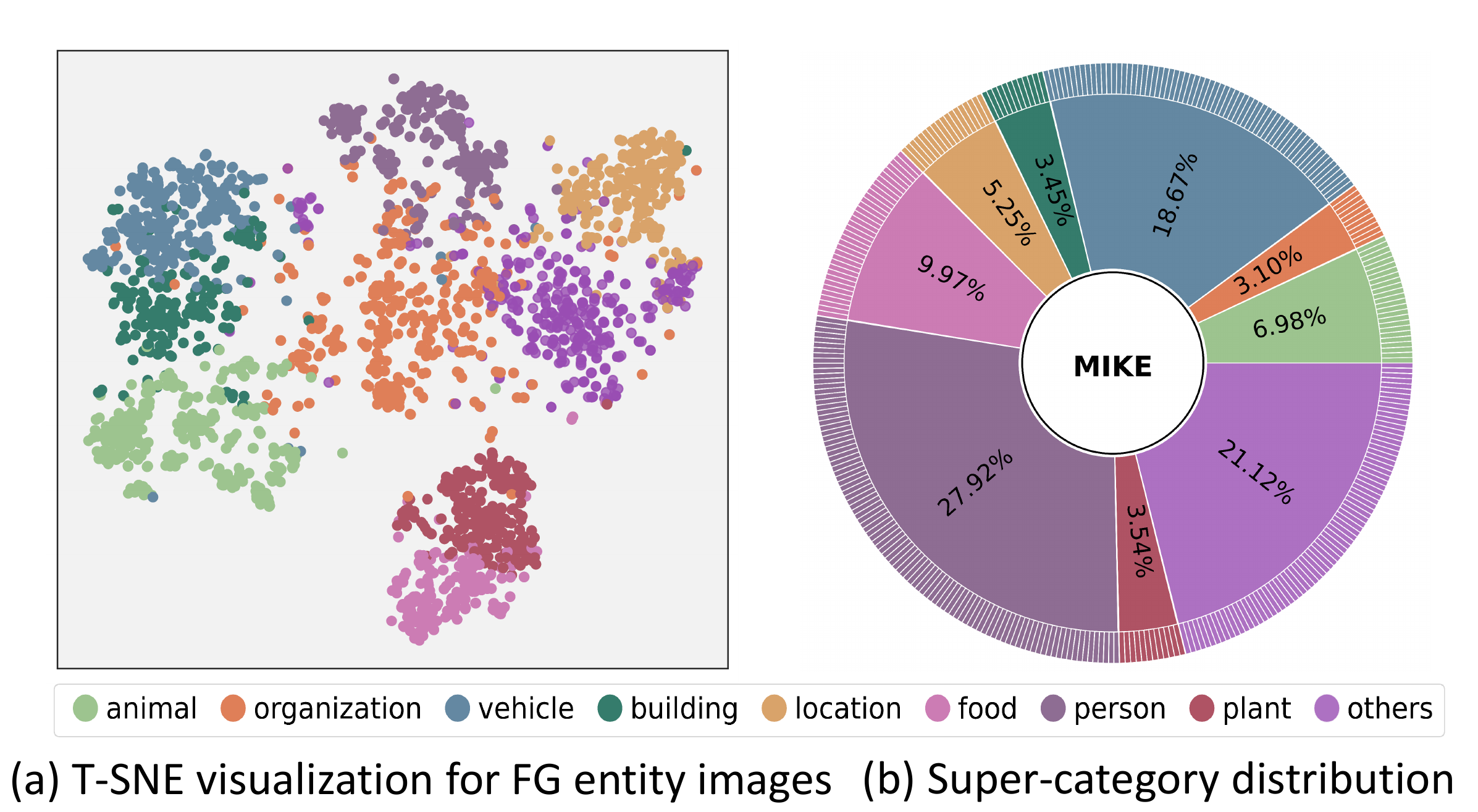}
\vspace{-0.6cm}
\caption{ Statistical analysis of MIKE. We utilize T-SNE to visualize the embeddings of FG entity images as can be seen in (a). The distribution of super-categories is shown in (b).}
\vspace{-0.15cm}
\label{figdata}
\end{figure}

\section{MIKE Benchmark}

\subsection{Collecting FG Entity Images}
\label{constructing}
\noindent \textbf{Collecting step.} To construct the FG multimodal entity dataset, we select 1500 FG entities from OVEN dataset \cite{DBLP:journals/corr/abs-2302-11154}, where each image is connected to a Wikipedia entity based on a text query. For each entity, we collect at least 5 different images from search engines like Google Search. Then we let 3 experienced annotators exclude the "dirty" images or entities. The collection rules are as follows:

\textbf{-Observable:}  This rule refers to the entities that could be described by images. We exclude words such as \emph{"1970s"} and \emph{"Love"} because they do not have descriptive visual features.

\textbf{-Specific:} FG entities are classified at an extremely detailed level. We exclude certain coarse-grained entities like \emph{"Africa"} and \emph{"Parent"} for their broad coverage and lack of distinctive visual features. 

\textbf{-Unambiguous:} An entity reference may correspond to multiple real-world entities, for instance, \emph{"Apple"} (fruit or company) and \emph{"Crane"} (machine or animal). We exclude these images from our dataset as they do not accurately depict the intended specific entities.

\textbf{-Unitary:}\label{unitary} An image may contain several entities, which may confuse MLLMs during the edit step. MLLMs do not know which is the target editing entity. We ensure that during the edit step, MLLMs could only see one editing entity in the image.

\noindent \textbf{Filtering step.} After collecting the images, we refine our dataset by filtering out FG entities already recognized by pre-trained MLLMs to construct a precise target set for editing. To facilitate this, we utilize prompts such as \emph{"Who is the character represented in this picture?"} to elicit specific FG entity names from MLLMs. To verify the pre-existence of entity knowledge within the models, we input all associated images for each entity into the MLLMs. An entity is considered pre-encoded in MLLMs if it is correctly identified from any of its images. Through this process, we determine that the final count of FG entities targeted for editing is 1,103.

\noindent \textbf{Data statistics.} The data statistics for collected entity images are summarized in Figure \ref{figdata}. We conduct a comprehensive count of all FG entities, categorizing them into 9 super-categories. In order to assess the quality of the collected images, we apply T-SNE \cite{van2008visualizing} to visualize the image embeddings, as depicted in Figure \ref{figdata} (a). The image embeddings are extracted using the Clip model \cite{DBLP:conf/icml/RadfordKHRGASAM21}. The visualization reveals that embeddings belonging to the same super-category are distinctly separated into compact clusters. It suggests that FG entities within each super-category share similar representations, which poses significant challenges for MKE. The distribution of super-categories can be observed in Figure \ref{figdata} (b). The super-category with the highest representation is \emph{person}, constituting 27.92\% of the entities. Intuitively, because each person represents a FG entity, \emph{person} presents more detailed and complex features compared to other super-categories.

\begin{figure}[t]
\centering
\includegraphics[width=\linewidth]{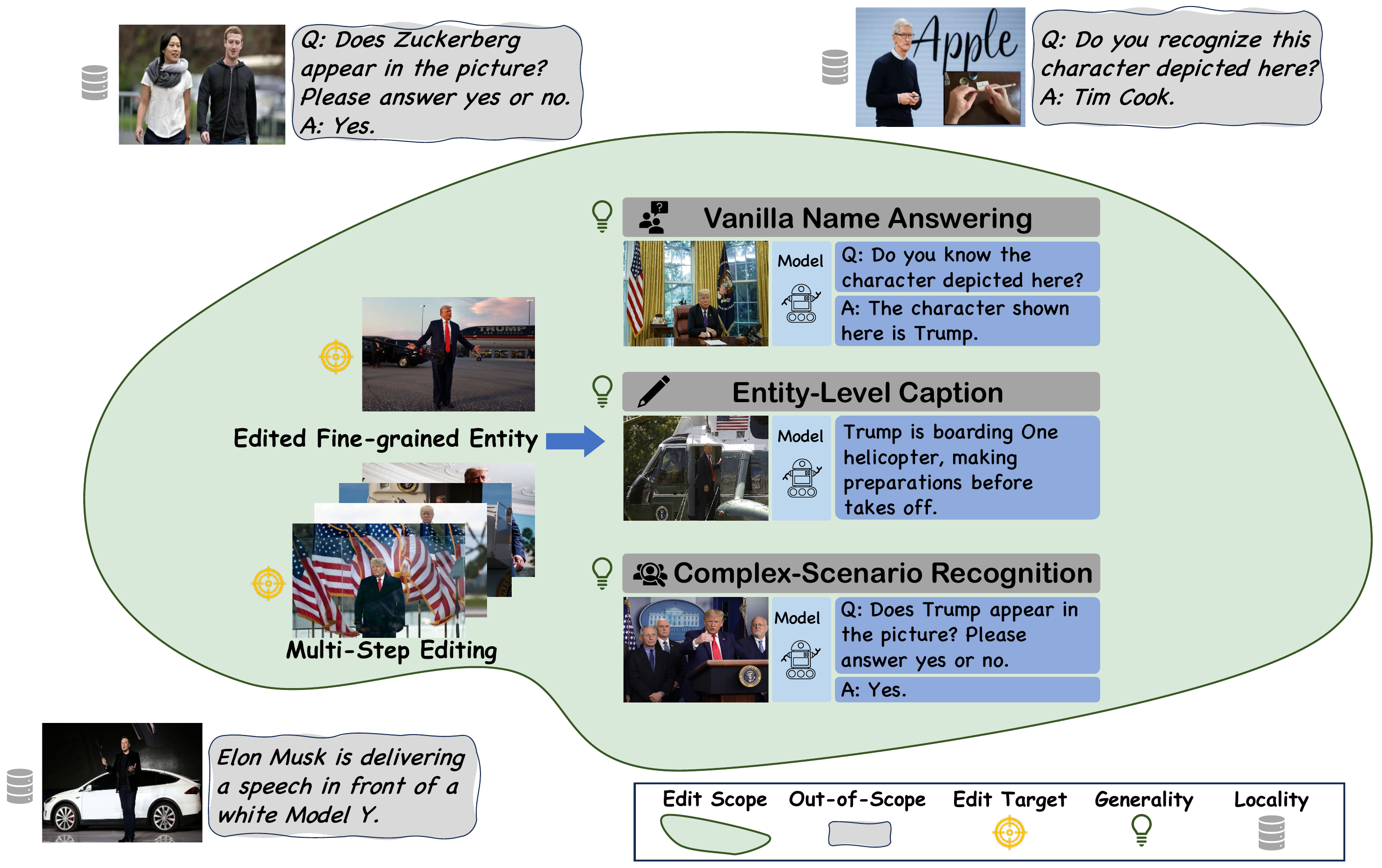}
\vspace{-0.6cm}
\caption{An example of editing \emph{Trump}.}
\vspace{-0.24cm}
\label{fig2}
\end{figure}


\subsection{Problem Formulation}\label{problem}

For a pre-trained MLLM $\mathcal{F}_\theta$ with parameter $\theta$, we have a target editing FG entity $\mathcal{E}$ which belongs to the constructed multimodal FG entity dataset $\mathcal{D}_m$. To ask MLLM for predicting entity related answer, we input a prompt $\mathcal{T}^i$ and an entity image $\mathcal{I}^i$ of $\mathcal{E}$. The original wrong output of $\mathcal{F}_\theta (\mathcal{T}^i,\mathcal{I}^i)$ is $\hat{y}^i$, where $\mathcal{E}\notin  \hat{y}^i$. To revise the incorrect answers and improve the recognition of FG entity knowledge, a multimodal knowledge editing method is utilized to edit $\mathcal{F}_\theta$. After the edit step, MLLM is optimized to $\mathcal{F}_{\widetilde{\theta}}$ with parameter $\widetilde{\theta}$. The ground truth is $y^i$, where $\mathcal{E}\in  y^i$. Inspired by MMEdit \cite{DBLP:conf/emnlp/0008TL0WC023}, we have three principles to guide the edit direction:

\textbf{-Reliability :} The goal of \emph{Reliability} is to modify the answer generated by MLLM from $\hat{y}^i$ to $y^i$.  The formula for the Reliability is structured as follows:
\begin{equation}
\label{for66}
    \mathcal{F}_{\widetilde{\theta}} (\mathcal{T}^i,\mathcal{I}^i)=y^i.
\end{equation}

\textbf{-Locality :} \emph{Locality} is to keep the prediction unchanged of out-of-scope entities. Following MMEdit, we split Locality into two parts, \emph{Text Locality} and \emph{Image Locality}. For Text Locality, the prediction of text-only input should be unchanged before and after the edit step. We use NQ Dataset $\mathcal{D}_t$ \cite{DBLP:journals/tacl/KwiatkowskiPRCP19} which could be regarded as out-of-scope examples to evaluate Text Locality:
\begin{equation}
\label{for66}
    \mathcal{F}_{\widetilde{\theta}} (\mathcal{X})=\mathcal{F}_\theta(\mathcal{X}), \mathcal{X} \in \mathcal{D}_t,
\end{equation}
where $\mathcal{X}$ is one of the examples in $\mathcal{D}_t$. 

In our dataset, as each entity represents a piece of independent FG knowledge, it could ensure that any other entity in our target editing set is the out-of-scope example of the target editing entity. Therefore, we randomly select the image $\mathcal{I}^*$ and question prompt $\mathcal{T}^*$ of another entity $\mathcal{E}^*$  as the example of Image Locality. Image Locality could be formulated as follows:
\begin{equation}
\label{for66}
    \mathcal{F}_{\widetilde{\theta}} (\mathcal{T}^*, \mathcal{I}^*)=\mathcal{F}_\theta(\mathcal{T}^*, \mathcal{I}^*).
\end{equation}

\textbf{-Generality :} To avoid overfitting, \emph{Generality} needs to be evaluated using the in-scope examples of $\mathcal{E}$ after the edit step. Similar to Locality, Generality is also split into \emph{Text Generality} and \emph{Image Generality}. For Text Generality, we utilize a rephrased prompt $\mathcal{T}^+$ of $\mathcal{T}^i$ as the in-scope prompt and the same image $\mathcal{I}^i$ with editing image. The objective of Text Generality is formulated as follows:
\begin{equation}
\label{for66}
    \mathcal{F}_{\widetilde{\theta}} (\mathcal{T}^+, \mathcal{I}^i)=\mathcal{F}_\theta(\mathcal{T}^i, \mathcal{I}^i).
\end{equation}

For Image Generality, MMEdit generates a new image using text-to-image tools with the same caption to reconstruct similar semantics. Different from MMEdit, we focus on the FG entity knowledge rather than global image content. We choose another image $\mathcal{I}^j$ of $\mathcal{E}$ as the example of Image Generality. Moreover, we create diverse tasks with corresponding prompts $\mathcal{T}^j$ to evaluate MKE methods as stated in Section \ref{sec:3.2}. Image Generality is defined as follows:
\begin{equation}
\label{for66}
    \mathcal{F}_{\widetilde{\theta}} (\mathcal{T}^j,\mathcal{I}^j)=y^j, i\ne j, 
\end{equation}
where $y^j$ is the ground truth of Image Generality.

\subsection{FG Multimodal Entity Tasks}
\label{sec:3.2}
We pose three FG entities oriented tasks over the collected images to form a benchmark as shown in Figure \ref{fig2}. Recent research \cite{DBLP:journals/tmlr/WeiTBRZBYBZMCHVLDF22,DBLP:journals/corr/abs-2311-00237} has revealed the emergent abilities in MLLMs, where MLLMs could exhibit surprising new capabilities via VQA interface. Inspired by the emergent abilities and many existing mainstream tasks, our designed tasks are tailored to test various aspects of an MLLM’s ability to recognize and interpret FG entities within multimodal contexts.

\subsubsection{Vanilla Name Answering}

\textbf{Motivation :} The core ability of FG multimodal entity knowledge editing lies in accurately identifying and naming entities for another image of the same entity after the edit step. Vanilla Name Answering (VNA) task simulates basic yet essential real-world applications like Multimodal Entity Linking \cite{DBLP:conf/acl/WangLZZPHMWWCXN23,DBLP:conf/sigir/WangWC22}, where precise entity identification is crucial.

\textbf{Details :} After the edit step, MLLMs are presented with images $\mathcal{I}^j$ containing target editing entities and are required to provide the short, precise name of the entity. To meet the condition of Image Generality, $\mathcal{I}^j$ is another image which is not used in the edit step. The prompt $\mathcal{T}^j$ is to instruct MLLMs to answer the short name of the FG entity such as \emph{"Question: Do you know the identity of the character depicted here? Short answer:"}. 

\subsubsection{Entity-Level Caption}
\textbf{Motivation :} Entity-Level Caption (ELC) task pushes MLLMs beyond mere recognition. In this task, MLLMs are tasked with creating captions for images that detail the scene and precisely identify and name the entities shown. This task draws inspiration from the emerging field of Entity-aware Captioning \cite{DBLP:conf/aaai/NguyenBMGK23,DBLP:conf/emnlp/ZhangGP23}. The Entity-aware Caption task typically requires additional background knowledge from the associated article to extract the FG entity name. In contrast, due to the knowledge already encoded in MLLMs through knowledge editing, our task eliminates the need for supplemental information. For example, MLLMs might directly generate a caption for a news image saying, \emph{"Trump is boarding One helicopter, making preparations before takes off
"}, providing a detailed narrative.


\textbf{Details :} In this task, MLLMs must create captions for images that describe the general scene while specifically naming the entities present. For the Image Generality image $\mathcal{I}^j$, we first generate the ground truth of ELC using LLaVA \cite{DBLP:journals/corr/abs-2304-08485}, a strong MLLM. To generate the caption containing the FG entity name, we carefully design an adaptive prompt to guide the MLLM to output the expected caption. Specifically, the prompt is \emph{"This is a \_. Please write a caption of the picture in a sentence. The caption must contain the word \_."}, where the blank space is filled by the FG entity name. In such way, each image $\mathcal{I}^j$ could be provided with its specific caption containing FG entity name to evaluate Image Generality. During Image Generality process, the prompt $\mathcal{T}^j$ is \emph{"Please write a caption of the picture in a sentence. The caption must contain the fine-grained entity names. Please include all fine-grained entity names as much as possible."}

\subsubsection{Complex-Scenario Recognition}

\textbf{Motivation :} The third task, Complex-Scenario Recognition (CSR), tests the MLLM’s performance in more complex scenarios where multiple entities are in an image. This task is inspired by Object Detection \cite{DBLP:journals/pieee/ZouCSGY23}, where the model needs to detect the pre-defined object surrounded by many other objects in the image. For our task setting, MLLMs need to correctly identify the edited FG entity, even when it is surrounded by multiple entities. This task is crucial for assessing the MLLM's ability to distinguish and focus on specific entities within crowded or complex scenes, a common challenge in real-world applications. For instance, MLLMs might be required to identify a known individual, such as \emph{"Does Trump appear in the picture?"} amidst a multitude of other entities.

\textbf{Details :} MLLMs are confronted with images featuring multiple entities, with the requirement to identify a specific edited entity among them. To set up challenging scenarios, we reserve images with complex contexts containing multiple entities during the initial collection of FG entity images, as mentioned in Section \ref{unitary}. These complex images do not go through the edit step but serve as the images $\mathcal{I}^j$ for CSR task. To give a more challenging setup, we employ a random seed when choosing $\mathcal{I}^j$ and constructing $\mathcal{T}^j$. The prompt $\mathcal{T}^j$ is \emph{"Does \_ appear in the picture? Please answer yes or no."}, where the blank space is randomly filled by edited entity name or another entity name. Likewise, the $\mathcal{I}^j$ is randomly chosen from the complex-scenario image of editing entity or another entity. If $\mathcal{T}^j$ and $\mathcal{I}^j$ are coreferential, the ground truth is yes, otherwise it is no.



\subsection{Multi-Step Editing}
\label{few}
 Multi-Step Editing examines the MLLMs' adaptability and learning efficiency which extends the normal knowledge editing form. In our form, MLLMs are evaluated on their performance in the above three tasks (VNA, ELC, and CSR) after editing 2-4 entity images. Multi-Step Editing is inspired by the Personalizing Text-to-Image Generation  \cite{DBLP:journals/corr/abs-2209-12330,DBLP:conf/siggrapha/ZengCXK23}, where Textual Inversion method \cite{DBLP:conf/iclr/GalAAPBCC23} utilizes 3-5 images to find the embedding space of a specific entity. After Multi-step images training, the model could freely generate the personalizing images and maintain existing abilities. Intuitively, this task and FG entity knowledge editing seem to be two parallel tasks with opposite data flows (One is text-to-image and another is image\&text-to-text). To this end, we wonder how many images do MLLMs need to edit an FG entity. Our task is designed to measure how quickly and effectively MLLMs can adapt to new FG entity knowledge and apply it across different tasks by Multi-Step Editing.

As stated in Section \ref{constructing}, we collect at least 5 images for each FG entity. Reserving an image for the Image Generality task, we test the performance on the above tasks by editing 2-4 images of the target editing FG entity during the edit step. After the edit step, we evaluate the above three tasks for each number of editing images.

\section{Experiments}
\label{exp}

\subsection{Evaluation Setup}
\noindent \textbf{MLLMs.} To evaluate MIKE benchmark, we conduct experiments on two MLLMs.    

\textbullet\ BLIP-2 \cite{DBLP:conf/icml/0008LSH23}: It consists of pre-trained visual encoders and text encoders with frozen parameters. BLIP-2 proposes a trainable Q-Former to act as a bottleneck between visual encoders and text encoders. Q-Former is a lightweight Transformer composed of a set of learnable Query vectors. 

\textbullet\ MiniGPT-4 \cite{DBLP:journals/corr/abs-2304-10592}: MiniGPT-4 aims to align the visual information from the pre-trained visual encoder with the Large Language Model. Specifically, Vicuna is used as a language decoder, which is based on LLaMA. For visual perception, MiniGPT-4 utilizes ViT backbone and pre-trained Q-Former, which are the same with BLIP-2.

\noindent \textbf{Baselines.} Following MMEdit, we test all multimodal knowledge editing methods incorporated in EasyEdit \cite{DBLP:journals/corr/abs-2308-07269} toolkit to conduct experiments. 

\textbullet\ MEND \cite{DBLP:conf/iclr/MitchellLBFM22}:  MEND trains lightweight model editor networks with the ability to generate edits to the weights of a pre-trained model. These edits are produced based on the standard fine-tuning gradient of a provided correction. MEND leverages the gradient as an information-rich starting point for the editing process.

\textbullet\ SERAC \cite{DBLP:conf/icml/MitchellLBMF22}: SERAC is composed of a scope classifier, a base model and a counterfactual model. The original model is no longer updated with parameters. The counterfactual model is a patch model to store new knowledge. Finally, a scope classifier is used to judge whether updated knowledge is needed. Then the classifier chooses to route to patch model or original model.

\textbullet\ IKE \cite{DBLP:conf/emnlp/ZhengLDFWXC23}: This method realizes knowledge editing by adding extra prompts in input. By studying several demonstrations, the edited models could update new facts without training. 

\noindent \textbf{Metrics.}  Different from MMEdit which directly uses token-level editing accuracy, we employ an entity-oriented metric. As many entity names are composed of two or more tokens, only one token recognized is regarded as a failure editing. To this end, for tokens of entity names, we employ entity exact match accuracy. The overall accuracy denoted $\mathcal{A} $ is formulated as follows:
\begin{equation}
\label{for66}
  \mathcal{A}= \frac{\mathds{1} \left [ \mathcal{F}_{\widetilde{\theta}} (\mathcal{T},\mathcal{I})=y \right ]+\mathds{1} \left [ \mathcal{E}\in \mathcal{F}_{\widetilde{\theta}} (\mathcal{T},\mathcal{I}) \right ]}{2}  ,
\end{equation}
where $\mathds{1}\left [ \cdot \right ]$ is the indicator function returning 1. The first half focuses on the token-level match, while another concerns the entity-level match.

\begin{table*}[t]
\renewcommand{\arraystretch}{0.95}
\renewcommand{\ttdefault}{pcr}

\centering
\scalebox{0.79}{
\begin{tabular}{c ccc ccc ccc ccc ccc}
\toprule
 \multirow{2}{*}{\textbf{Method}} 
 & \multicolumn{5}{c}{\textbf{Vanilla Name Answering} } & \multicolumn{5}{c}{\textbf{Entity-Level Caption } } & \multicolumn{5}{c}{\textbf{Complex-Scenario Recognition} } \\ 

\cmidrule(r){2-6} \cmidrule(r){7-11} \cmidrule(r){12-16} 
 
  &\textbf{R} &\textbf{I-G}&\textbf{I-L}&\textbf{T-G} &\textbf{T-L}&\textbf{R} &\textbf{I-G}&\textbf{I-L}&\textbf{T-G} &\textbf{T-L}&\textbf{R} &\textbf{I-G}&\textbf{I-L}&\textbf{T-G} &\textbf{T-L}\\

\midrule

\multicolumn{16}{c}{\textbf{BLIP-2 OPT} }\\

\midrule

   MEND\textsubscript{opt 2.7B} & 87.2 & 67.3 & 35.1 & 88.6 & 94.1
 & \multicolumn{1}{|l}{48.7} & 16.7 & 37.5 & 71.0 & 97.9 & \multicolumn{1}{|l}{80.6} & 50.3& 26.9 & 81.7 & 85.2\\
    MEND\textsubscript{opt 6.7B} & 85.8 & 70.7 & 36.0 & 85.1 & 97.3
 & \multicolumn{1}{|l}{50.3} & 13.1 & 42.8 & 71.3 & 95.8 & \multicolumn{1}{|l}{83.2} & \cellcolor[rgb]{0.9255, 0.8196, 0.708}51.7& 22.0 & 80.0 & 81.3\\
 \midrule

   SERAC\textsubscript{opt 2.7B} & 87.8 & 72.5 & 18.3 & 90.8 & \cellcolor[rgb]{0.8509, 0.6392, 0.416}100.0 &\multicolumn{1}{|l} {82.4} & 69.8 & 19.2 & 83.9 & \cellcolor[rgb]{0.8509, 0.6392, 0.416}99.9 &\multicolumn{1}{|l}{85.4} & \cellcolor[rgb]{0.8509, 0.6392, 0.416}100.0& 23.2 & 87.1 & \cellcolor[rgb]{0.8509, 0.6392, 0.416}100.0\\
       SERAC\textsubscript{opt 6.7B} & 89.3 & 69.2 & 13.1 & 94.6 & \cellcolor[rgb]{0.8509, 0.6392, 0.416}100.0
 & \multicolumn{1}{|l}{79.2} & \cellcolor[rgb]{0.9255, 0.8196, 0.708}72.2 & 17.1 & \cellcolor[rgb]{0.9255, 0.8196, 0.708}85.9 & \cellcolor[rgb]{0.9255, 0.8196, 0.708}99.7 & \multicolumn{1}{|l}{84.5} & \cellcolor[rgb]{0.8509, 0.6392, 0.416}100.0& 21.5 & 81.7 & \cellcolor[rgb]{0.8509, 0.6392, 0.416}100.0\\
 \midrule
      IKE\textsubscript{opt 2.7B} & 94.6 & \cellcolor[rgb]{0.8509, 0.6392, 0.416}94.2 & \cellcolor[rgb]{0.9255, 0.8196, 0.708}88.7 & \cellcolor[rgb]{0.8509, 0.6392, 0.416}96.8 & 99.1 & \multicolumn{1}{|l}{83.6} & 8.8 & \cellcolor[rgb]{0.8509, 0.6392, 0.416}85.4 & 33.1 & 82.6 &\multicolumn{1}{|l}{86.2} & 28.2& \cellcolor[rgb]{0.9255, 0.8196, 0.708}87.3 & 99.1 & \cellcolor[rgb]{0.8509, 0.6392, 0.416}100.0\\
          IKE\textsubscript{opt 6.7B} & \cellcolor[rgb]{0.9255, 0.8196, 0.708}96.1 & 92.8 & \cellcolor[rgb]{0.8509, 0.6392, 0.416}90.5 & 94.3 & \cellcolor[rgb]{0.9255, 0.8196, 0.708}99.6
 & \multicolumn{1}{|l}{\cellcolor[rgb]{0.8509, 0.6392, 0.416}{86.8}} & 5.4 & \cellcolor[rgb]{0.9255, 0.8196, 0.708}82.8 & 31.0 & 77.4 & \multicolumn{1}{|l}{84.1} & 23.1& \cellcolor[rgb]{0.8509, 0.6392, 0.416}89.4 & \cellcolor[rgb]{0.9255, 0.8196, 0.708}99.4 & \cellcolor[rgb]{0.8509, 0.6392, 0.416}100.0\\
 
\midrule

\multicolumn{16}{c}{\textbf{MiniGPT-4}\textsubscript{7.3B} }\\
      
\midrule

   MEND & 88.4 & 69.4 & 32.5 & 89.4 & 96.4 & \multicolumn{1}{|l}{54.2} & 17.4 & 34.1 & 68.4 & 98.8 & \multicolumn{1}{|l}{78.4} & 49.6& 20.8 & 85.6 & \cellcolor[rgb]{0.9255, 0.8196, 0.708}87.9\\

   SERAC & 91.6 & 72.3 & 11.6 & 93.0 & \cellcolor[rgb]{0.8509, 0.6392, 0.416}100.0 &\multicolumn{1}{|l} {80.3} & \cellcolor[rgb]{0.8509, 0.6392, 0.416}74.6 & 13.7 & \cellcolor[rgb]{0.8509, 0.6392, 0.416}86.2 & 99.3 &\multicolumn{1}{|l}{\cellcolor[rgb]{0.9255, 0.8196, 0.708}{87.5}} & \cellcolor[rgb]{0.8509, 0.6392, 0.416}100.0& 18.5 & 89.5 & \cellcolor[rgb]{0.8509, 0.6392, 0.416}100.0\\
      IKE & \cellcolor[rgb]{0.8509, 0.6392, 0.416}97.5 & \cellcolor[rgb]{0.9255, 0.8196, 0.708}93.1 & 86.3 & \cellcolor[rgb]{0.9255, 0.8196, 0.708}95.7 & 98.4 & \multicolumn{1}{|l}{\cellcolor[rgb]{0.9255, 0.8196, 0.708}{84.6}} & 9.0 & 79.9 & 36.4 & 81.5 &\multicolumn{1}{|l}{\cellcolor[rgb]{0.8509, 0.6392, 0.416}{88.4}} & 26.6& 84.6 & \cellcolor[rgb]{0.8509, 0.6392, 0.416}99.6 & \cellcolor[rgb]{0.8509, 0.6392, 0.416}100.0\\

\bottomrule

\end{tabular}}

\caption{Overall results on three tasks. `\textit{R}', `\textit{I-G}',`\textit{I-L}',`\textit{T-G}' and `\textit{T-L}' represent the \textbf{Reliability}, \textbf{Image Generality}, \textbf{Image locality}, \textbf{Text Generality} and \textbf{Text Locality} respectively.}
\vspace{-0.3cm}
\label{tab:all}
\end{table*}

\subsection{Results \& Analysis}

We report the results of VNA, ELC and CSR in Table \ref{tab:all}. Our main observations are summarized as follows:

\textbf{\textit{(i)}} Our first observation is that each editing method exhibits specific weaknesses. IKE stands out in VNA, delivering the highest performance across all aspects but showing lower accuracy in Image Generality and Text Generality for ELC. This discrepancy might stem from the nature of VNA, where predictions are brief and closely aligned with the editing labels, allowing IKE to excel in this simpler question-answer format without additional MLLM training. 
Conversely, SERAC demonstrates high Image Generality accuracy across all tasks, showcasing the robustness of its editing approach. Nonetheless, it underperforms in Image Locality, potentially due to its classifier misidentifying out-of-scope examples. This issue likely arises because SERAC's counterfact model is tailored to only restore the knowledge of in-scope data, rendering it less adaptable to out-of-scope queries. Overall, these findings demonstrate that current editing methods were unable to thoroughly address all aspects due to the complexities of our task. 

 \textbf{\textit{(ii)}} In evaluating Image Generality across the three tasks, it's evident that all editing methods show their weakest performance on the ELC task. Specifically, for BLIP-2 OPT's ELC, MEND records a notably low accuracy of 16.7\%, a stark contrast to its 67.3\% on VNA and 50.3\% on CSR. Similarly, SERAC achieves 100\% in Image Generality for CSR but drops to 69.8\% for ELC. This trend suggests that the ELC task, which requires simultaneous recognition of FG entities and understanding of the overall image content, poses a significant challenge to MLLMs. The disparity in the level of comprehension highlights the ELC task as the biggest challenge for all editing methods.

\textbf{\textit{(iii)}} Table \ref{tab:all} shows that different Image Generality tasks affect other aspects. MEND exhibits poorer performance in Reliability and Text Generality on the ELC task compared to the other two tasks. Moreover, MEND, SERAC, and IKE all achieve their highest Reliability scores on the VNA task. A contributing factor to this pattern might be that each editing method must calculate gradients for Reliability, Generality, and Locality during the editing process. This joint calculation leads MLLMs to extract diverse semantic features for different Image Generality tasks, impacting other aspects through the backpropagation process.

\textbf{\textit{(iv)}} We also observe that model sizes are not that critical. Although MiniGPT-4 is much larger than BLIP-2 OPT, the gap in performance is not obvious. In some aspects, BLIP-2 even performs better than MiniGPT-4. For instance, each method achieves more Image Locality accuracy on VNA using BLIP-2 than MiniGPT-4. In addition, we leverage BLIP-2 OPT 2.7B and 6.7B as our baselines. The results show that they perform competitively. The reason is perhaps that knowledge editing does not need to encode much knowledge into MLLMs. Thus the demands on model size are not so great.

\begin{figure*}[t]
\centering
\includegraphics[width=0.99\textwidth]{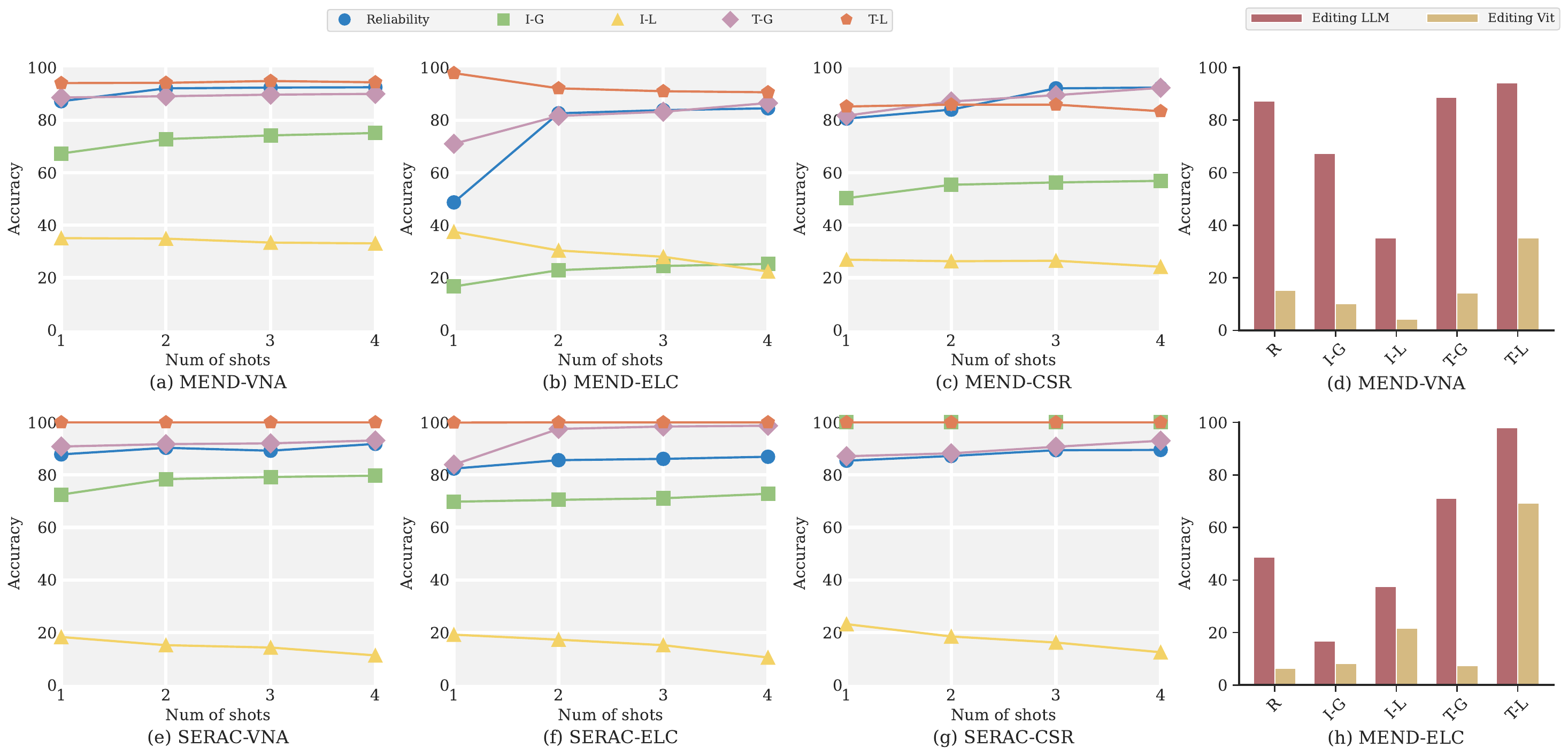}
\vspace{-0.1cm}
\caption{ The effects of Multi-Step Editing on MEND (a) - (c) and SERAC (e) - (g).  (d) and (h) show the results of editing LLM and Vit using MEND. We report the \textbf{Reliability}, \textbf{Image Generality}, \textbf{Image locality}, \textbf{Text Generality} and \textbf{Text Locality}.}
\vspace{-0.35cm}
\label{fig5}
\end{figure*}

\subsection{Effects of Multi-Step Editing}

Figure \ref{fig5} shows the impacts of Multi-Step Editing. From each experiment, we could observe that the Reliability, Image Generality and Text Generality could be improved by adding the editing images of FG entities. Among them, the most improvement is Reliability, as evidenced in \textbf{MEND-ELC} from 48.7\% to 81.2\%. This demonstrates that the mapping between the visual appearance and the FG text name could be refined by Multi-Step Editing. It is noted that Image Generality accuracy is significantly boosted from 80.6\% to 92.5\% as can be seen in \textbf{MEND-CSR}. A high Image Generality accuracy could prove that the multimodal features of FG entities are greatly encoded into MLLMs. We could find that the Text Generality is slightly improved compared with Reliability and Image Generality. The reason may be that more editing images do not have the information to improve textual features.

\begin{table}[t]
\renewcommand{\arraystretch}{0.95}
\renewcommand{\ttdefault}{pcr}

\centering
\scalebox{0.9}{
\begin{tabular}{c ccc cc}
\toprule
\textbf{AUG method}
 
  &\textbf{R} &\textbf{I-G}&\textbf{I-L}&\textbf{T-G} &\textbf{T-L}\\

\midrule

  w/o AUG & 87.2 & 67.3 & \cellcolor[rgb]{0.8509, 0.6392, 0.416}35.1 & 88.6 & \cellcolor[rgb]{0.8509, 0.6392, 0.416}94.1\\
    Vertical Flip & \cellcolor[rgb]{0.9255, 0.8196, 0.708}87.4 & 72.3 & 28.4 & \cellcolor[rgb]{0.8509, 0.6392, 0.416}93.2 & 92.0\\
 
   Horizontal Flip & 85.4 & 69.4 & 30.5 & 88.1 & 90.5 \\
       Random Noise & \cellcolor[rgb]{0.8509, 0.6392, 0.416}92.5 & \cellcolor[rgb]{0.8509, 0.6392, 0.416}75.5 & \cellcolor[rgb]{0.9255, 0.8196, 0.708}33.4 & \cellcolor[rgb]{0.9255, 0.8196, 0.708}92.2 & \cellcolor[rgb]{0.9255, 0.8196, 0.708}92.7
 \\
 
      Color Jitter & 87.3 & \cellcolor[rgb]{0.9255, 0.8196, 0.708}73.6 & 31.5 & 90.9 & 91.8 \\

\bottomrule

\end{tabular}}

\caption{Results of applying augmentations to images. \textbf{w/o AUG} means the images are not equipped with augmentations.}
\vspace{-0.3cm}
\label{tab:aug}
\end{table}

In addition, we observe that two-step editing brings the most improvement. The changes of three-step editing and four-step editing are relatively smaller than two-step editing. It means that after four-step editing the accuracy tends to converge gradually. Jointly considering the statement in Section \ref{few},  the phenomenon proves that a multimodal FG entity could be edited into MLLMs admirably with 3-4 images. The reason may be that only one entity image could not cover all the features of the FG entities, while 3-4 images could encode most features of the entities.  Further increasing the number of edited images does not lead to a significant improvement in accuracy. We also notice that the Text Locality and Image Locality accuracy decreases or is unchanged after Multi-Step Editing, especially the Image Locality. With the increase in editing images, the degree of decline becomes bigger.



\subsection{Comparison with Editing ViT}

As recognizing FG multimodal entity requires a strong discriminative ability of visual features extracted by MLLMs, we compare the form of editing LLM with editing ViT which is the visual encoder of BLIP-2. The experiment results of VNA and ELC are shown in Figure \ref{fig5}. Intuitively, editing ViT could directly help MLLMs understand the visual appearance of FG entities. However, it could be observed that compared to editing the layers of LLM, every aspect accuracy of editing ViT is far behind. It is perhaps that even though the visual encoder is refined, the mapping module Q-Former keeps frozen. The presence of the frozen Q-Former restricts the joint understanding of both LLM and ViT, leading to incorrect predictions by LLM.

\subsection{Impacts of Image Augmentations}


We explored the impacts of image augmentations during the editing process on performance improvement. We applied the MEND method to the VNA task. Our experiments examined four augmentation strategies: \textbf{Vertical Flip}, \textbf{Horizontal Flip}, \textbf{Random Noise}, and \textbf{Color Jitter}, which are usually utilized in Computer Vision tasks such as Image Classification \cite{DBLP:conf/iccv/ChenFP21}, Object Detection \cite{DBLP:journals/pieee/ZouCSGY23}, etc. As shown in Table \ref{tab:aug}, we observe that: (i) all augmentation methods enhance the Image Generality score; (ii) Random Noise notably increases both Image Generality and Reliability; (iii) images without augmentations achieve the highest Locality scores.

\section{Conclusion}

We present MIKE: a benchmark which aims to edit FG multimodal entity knowledge into MLLMs. Our dataset contains a large and diverse set of FG entities. We introduce three challenging tasks, VNA, ELC and CSR to evaluate the generality of editing methods. Finally, we present a new form of Multi-Step Editing compared with normal Knowledge Editing. For future work,  we would try to extend this work mainly in following aspects: (i) continually collecting diverse FG multimodal entities;  (ii) evaluating more editing methods on MIKE; (iii) proposing a new editing method to more effectively edit FG multimodal entities into MLLMs.

\section*{Limitations}
The main limitations of our work are related to the editing methods. The EasyEdit toolkit we utilized does not encompass all existing editing methods, so we only evaluated the MEND, SERAC, and IKE editing methods. Another limitation pertains to the models. Due to limited computing resources, we only tested BLIP-2 OPT and MiniGPT-4.

\bibliography{anthology,custom}

\begin{thebibliography}{45}
\expandafter\ifx\csname natexlab\endcsname\relax\def\natexlab#1{#1}\fi

\bibitem[{Alayrac et~al.(2022)Alayrac, Donahue, Luc, Miech, Barr, Hasson, Lenc, Mensch, Millican, Reynolds, Ring, Rutherford, Cabi, Han, Gong, Samangooei, Monteiro, Menick, Borgeaud, Brock, Nematzadeh, Sharifzadeh, Binkowski, Barreira, Vinyals, Zisserman, and Simonyan}]{DBLP:conf/nips/AlayracDLMBHLMM22}
Jean{-}Baptiste Alayrac, Jeff Donahue, Pauline Luc, Antoine Miech, Iain Barr, Yana Hasson, Karel Lenc, Arthur Mensch, Katherine Millican, Malcolm Reynolds, Roman Ring, Eliza Rutherford, Serkan Cabi, Tengda Han, Zhitao Gong, Sina Samangooei, Marianne Monteiro, Jacob~L. Menick, Sebastian Borgeaud, Andy Brock, Aida Nematzadeh, Sahand Sharifzadeh, Mikolaj Binkowski, Ricardo Barreira, Oriol Vinyals, Andrew Zisserman, and Kar{\'{e}}n Simonyan. 2022.
\newblock Flamingo: a visual language model for few-shot learning.
\newblock In \emph{NeurIPS}.

\bibitem[{Bai et~al.(2023)Bai, Bai, Yang, Wang, Tan, Wang, Lin, Zhou, and Zhou}]{DBLP:journals/corr/abs-2308-12966}
Jinze Bai, Shuai Bai, Shusheng Yang, Shijie Wang, Sinan Tan, Peng Wang, Junyang Lin, Chang Zhou, and Jingren Zhou. 2023.
\newblock Qwen-vl: {A} frontier large vision-language model with versatile abilities.
\newblock \emph{CoRR}, abs/2308.12966.

\bibitem[{Cao et~al.(2021)Cao, Aziz, and Titov}]{DBLP:conf/emnlp/CaoAT21}
Nicola~De Cao, Wilker Aziz, and Ivan Titov. 2021.
\newblock Editing factual knowledge in language models.
\newblock In \emph{{EMNLP} {(1)}}, pages 6491--6506. Association for Computational Linguistics.

\bibitem[{Chen et~al.(2021)Chen, Fan, and Panda}]{DBLP:conf/iccv/ChenFP21}
Chun{-}Fu~(Richard) Chen, Quanfu Fan, and Rameswar Panda. 2021.
\newblock Crossvit: Cross-attention multi-scale vision transformer for image classification.
\newblock In \emph{{ICCV}}, pages 347--356. {IEEE}.

\bibitem[{Chen et~al.(2023)Chen, Hu, Luan, Sun, Changpinyo, Ritter, and Chang}]{DBLP:conf/emnlp/ChenHLSCRC23}
Yang Chen, Hexiang Hu, Yi~Luan, Haitian Sun, Soravit Changpinyo, Alan Ritter, and Ming{-}Wei Chang. 2023.
\newblock Can pre-trained vision and language models answer visual information-seeking questions?
\newblock In \emph{{EMNLP}}, pages 14948--14968. Association for Computational Linguistics.

\bibitem[{Cheng et~al.(2023)Cheng, Tian, Liu, Chen, Wang, Chen, and Zhang}]{DBLP:conf/emnlp/0008TL0WC023}
Siyuan Cheng, Bozhong Tian, Qingbin Liu, Xi~Chen, Yongheng Wang, Huajun Chen, and Ningyu Zhang. 2023.
\newblock Can we edit multimodal large language models?
\newblock In \emph{{EMNLP}}, pages 13877--13888. Association for Computational Linguistics.

\bibitem[{Devlin et~al.(2019)Devlin, Chang, Lee, and Toutanova}]{DBLP:conf/naacl/DevlinCLT19}
Jacob Devlin, Ming{-}Wei Chang, Kenton Lee, and Kristina Toutanova. 2019.
\newblock {BERT:} pre-training of deep bidirectional transformers for language understanding.
\newblock In \emph{{NAACL-HLT} {(1)}}, pages 4171--4186. Association for Computational Linguistics.

\bibitem[{Dong et~al.(2022)Dong, Dai, Song, Xu, Sui, and Li}]{DBLP:conf/emnlp/DongDSXSL22}
Qingxiu Dong, Damai Dai, Yifan Song, Jingjing Xu, Zhifang Sui, and Lei Li. 2022.
\newblock Calibrating factual knowledge in pretrained language models.
\newblock In \emph{{EMNLP} (Findings)}, pages 5937--5947. Association for Computational Linguistics.

\bibitem[{Gal et~al.(2023)Gal, Alaluf, Atzmon, Patashnik, Bermano, Chechik, and Cohen{-}Or}]{DBLP:conf/iclr/GalAAPBCC23}
Rinon Gal, Yuval Alaluf, Yuval Atzmon, Or~Patashnik, Amit~Haim Bermano, Gal Chechik, and Daniel Cohen{-}Or. 2023.
\newblock An image is worth one word: Personalizing text-to-image generation using textual inversion.
\newblock In \emph{{ICLR}}. OpenReview.net.

\bibitem[{Gallego(2022)}]{DBLP:journals/corr/abs-2209-12330}
V{\'{\i}}ctor Gallego. 2022.
\newblock Personalizing text-to-image generation via aesthetic gradients.
\newblock \emph{CoRR}, abs/2209.12330.

\bibitem[{Guo et~al.(2023)Guo, Liu, Ren, Grosz, Masi, and Liu}]{DBLP:conf/cvpr/GuoLRGM023}
Xiao Guo, Xiaohong Liu, Zhiyuan Ren, Steven Grosz, Iacopo Masi, and Xiaoming Liu. 2023.
\newblock Hierarchical fine-grained image forgery detection and localization.
\newblock In \emph{{CVPR}}, pages 3155--3165. {IEEE}.

\bibitem[{Hase et~al.(2023)Hase, Diab, Celikyilmaz, Li, Kozareva, Stoyanov, Bansal, and Iyer}]{DBLP:conf/eacl/HaseDCLKSBI23}
Peter Hase, Mona~T. Diab, Asli Celikyilmaz, Xian Li, Zornitsa Kozareva, Veselin Stoyanov, Mohit Bansal, and Srinivasan Iyer. 2023.
\newblock Methods for measuring, updating, and visualizing factual beliefs in language models.
\newblock In \emph{{EACL}}, pages 2706--2723. Association for Computational Linguistics.

\bibitem[{Hu et~al.(2023{\natexlab{a}})Hu, Luan, Chen, Khandelwal, Joshi, Lee, Toutanova, and Chang}]{DBLP:journals/corr/abs-2302-11154}
Hexiang Hu, Yi~Luan, Yang Chen, Urvashi Khandelwal, Mandar Joshi, Kenton Lee, Kristina Toutanova, and Ming{-}Wei Chang. 2023{\natexlab{a}}.
\newblock Open-domain visual entity recognition: Towards recognizing millions of wikipedia entities.
\newblock \emph{CoRR}, abs/2302.11154.

\bibitem[{Hu et~al.(2023{\natexlab{b}})Hu, Hua, Yang, Shi, Smith, and Luo}]{DBLP:journals/corr/abs-2211-09699}
Yushi Hu, Hang Hua, Zhengyuan Yang, Weijia Shi, Noah~A. Smith, and Jiebo Luo. 2023{\natexlab{b}}.
\newblock Promptcap: Prompt-guided task-aware image captioning.
\newblock In \emph{{ICCV}}.

\bibitem[{Khan et~al.(2023)Khan, Kumar, Schulter, Yu, Fu, and Chandraker}]{DBLP:conf/cvpr/0001KS00C23}
Zaid Khan, BG~Vijay Kumar, Samuel Schulter, Xiang Yu, Yun Fu, and Manmohan Chandraker. 2023.
\newblock {Q:} how to specialize large vision-language models to data-scarce {VQA} tasks? {A:} self-train on unlabeled images!
\newblock In \emph{{CVPR}}, pages 15005--15015. {IEEE}.

\bibitem[{Kwiatkowski et~al.(2019)Kwiatkowski, Palomaki, Redfield, Collins, Parikh, Alberti, Epstein, Polosukhin, Devlin, Lee, Toutanova, Jones, Kelcey, Chang, Dai, Uszkoreit, Le, and Petrov}]{DBLP:journals/tacl/KwiatkowskiPRCP19}
Tom Kwiatkowski, Jennimaria Palomaki, Olivia Redfield, Michael Collins, Ankur~P. Parikh, Chris Alberti, Danielle Epstein, Illia Polosukhin, Jacob Devlin, Kenton Lee, Kristina Toutanova, Llion Jones, Matthew Kelcey, Ming{-}Wei Chang, Andrew~M. Dai, Jakob Uszkoreit, Quoc Le, and Slav Petrov. 2019.
\newblock Natural questions: a benchmark for question answering research.
\newblock \emph{Trans. Assoc. Comput. Linguistics}, 7:452--466.

\bibitem[{Li et~al.(2023{\natexlab{a}})Li, Zhang, Chen, Wang, Yang, and Liu}]{DBLP:journals/corr/abs-2305-03726}
Bo~Li, Yuanhan Zhang, Liangyu Chen, Jinghao Wang, Jingkang Yang, and Ziwei Liu. 2023{\natexlab{a}}.
\newblock Otter: {A} multi-modal model with in-context instruction tuning.
\newblock \emph{CoRR}, abs/2305.03726.

\bibitem[{Li et~al.(2023{\natexlab{b}})Li, Li, Savarese, and Hoi}]{DBLP:conf/icml/0008LSH23}
Junnan Li, Dongxu Li, Silvio Savarese, and Steven C.~H. Hoi. 2023{\natexlab{b}}.
\newblock {BLIP-2:} bootstrapping language-image pre-training with frozen image encoders and large language models.
\newblock In \emph{{ICML}}, volume 202 of \emph{Proceedings of Machine Learning Research}, pages 19730--19742. {PMLR}.

\bibitem[{Liu et~al.(2023)Liu, Li, Wu, and Lee}]{DBLP:journals/corr/abs-2304-08485}
Haotian Liu, Chunyuan Li, Qingyang Wu, and Yong~Jae Lee. 2023.
\newblock Visual instruction tuning.
\newblock \emph{CoRR}, abs/2304.08485.

\bibitem[{Meng et~al.(2022)Meng, Bau, Andonian, and Belinkov}]{DBLP:conf/nips/MengBAB22}
Kevin Meng, David Bau, Alex Andonian, and Yonatan Belinkov. 2022.
\newblock Locating and editing factual associations in {GPT}.
\newblock In \emph{NeurIPS}.

\bibitem[{Meng et~al.(2023)Meng, Sharma, Andonian, Belinkov, and Bau}]{DBLP:conf/iclr/MengSABB23}
Kevin Meng, Arnab~Sen Sharma, Alex~J. Andonian, Yonatan Belinkov, and David Bau. 2023.
\newblock Mass-editing memory in a transformer.
\newblock In \emph{{ICLR}}. OpenReview.net.

\bibitem[{Mitchell et~al.(2022{\natexlab{a}})Mitchell, Lin, Bosselut, Finn, and Manning}]{DBLP:conf/iclr/MitchellLBFM22}
Eric Mitchell, Charles Lin, Antoine Bosselut, Chelsea Finn, and Christopher~D. Manning. 2022{\natexlab{a}}.
\newblock Fast model editing at scale.
\newblock In \emph{{ICLR}}. OpenReview.net.

\bibitem[{Mitchell et~al.(2022{\natexlab{b}})Mitchell, Lin, Bosselut, Manning, and Finn}]{DBLP:conf/icml/MitchellLBMF22}
Eric Mitchell, Charles Lin, Antoine Bosselut, Christopher~D. Manning, and Chelsea Finn. 2022{\natexlab{b}}.
\newblock Memory-based model editing at scale.
\newblock In \emph{{ICML}}, volume 162 of \emph{Proceedings of Machine Learning Research}, pages 15817--15831. {PMLR}.

\bibitem[{Nguyen et~al.(2023)Nguyen, Biten, Mafla, G{\'{o}}mez, and Karatzas}]{DBLP:conf/aaai/NguyenBMGK23}
Khanh Nguyen, Ali~Furkan Biten, Andr{\'{e}}s Mafla, Llu{\'{\i}}s G{\'{o}}mez, and Dimosthenis Karatzas. 2023.
\newblock Show, interpret and tell: Entity-aware contextualised image captioning in wikipedia.
\newblock In \emph{{AAAI}}, pages 1940--1948. {AAAI} Press.

\bibitem[{Radford et~al.(2021)Radford, Kim, Hallacy, Ramesh, Goh, Agarwal, Sastry, Askell, Mishkin, Clark, Krueger, and Sutskever}]{DBLP:conf/icml/RadfordKHRGASAM21}
Alec Radford, Jong~Wook Kim, Chris Hallacy, Aditya Ramesh, Gabriel Goh, Sandhini Agarwal, Girish Sastry, Amanda Askell, Pamela Mishkin, Jack Clark, Gretchen Krueger, and Ilya Sutskever. 2021.
\newblock Learning transferable visual models from natural language supervision.
\newblock In \emph{{ICML}}, volume 139 of \emph{Proceedings of Machine Learning Research}, pages 8748--8763. {PMLR}.

\bibitem[{Raffel et~al.(2020)Raffel, Shazeer, Roberts, Lee, Narang, Matena, Zhou, Li, and Liu}]{DBLP:journals/jmlr/RaffelSRLNMZLL20}
Colin Raffel, Noam Shazeer, Adam Roberts, Katherine Lee, Sharan Narang, Michael Matena, Yanqi Zhou, Wei Li, and Peter~J. Liu. 2020.
\newblock Exploring the limits of transfer learning with a unified text-to-text transformer.
\newblock \emph{J. Mach. Learn. Res.}, 21:140:1--140:67.

\bibitem[{Ramos et~al.(2023)Ramos, Martins, Elliott, and Kementchedjhieva}]{DBLP:conf/cvpr/Ramos0EK23}
Rita Ramos, Bruno Martins, Desmond Elliott, and Yova Kementchedjhieva. 2023.
\newblock Smallcap: Lightweight image captioning prompted with retrieval augmentation.
\newblock In \emph{{CVPR}}, pages 2840--2849. {IEEE}.

\bibitem[{Tang et~al.(2023)Tang, Yang, and Chen}]{DBLP:conf/cvpr/TangYC23}
Zhenchao Tang, Hualin Yang, and Calvin~Yu{-}Chian Chen. 2023.
\newblock Weakly supervised posture mining for fine-grained classification.
\newblock In \emph{{CVPR}}, pages 23735--23744. {IEEE}.

\bibitem[{Tanno et~al.(2022)Tanno, Pradier, Nori, and Li}]{DBLP:conf/nips/TannoPNL22}
Ryutaro Tanno, Melanie~F. Pradier, Aditya~V. Nori, and Yingzhen Li. 2022.
\newblock Repairing neural networks by leaving the right past behind.
\newblock In \emph{NeurIPS}.

\bibitem[{Van~der Maaten and Hinton(2008)}]{van2008visualizing}
Laurens Van~der Maaten and Geoffrey Hinton. 2008.
\newblock Visualizing data using t-sne.
\newblock \emph{JMLR}.

\bibitem[{Wang et~al.(2022)Wang, Wu, and Chen}]{DBLP:conf/sigir/WangWC22}
Peng Wang, Jiangheng Wu, and Xiaohang Chen. 2022.
\newblock Multimodal entity linking with gated hierarchical fusion and contrastive training.
\newblock In \emph{{SIGIR}}, pages 938--948. {ACM}.

\bibitem[{Wang et~al.(2023{\natexlab{a}})Wang, Zhang, Xie, Yao, Tian, Wang, Xi, Cheng, Liu, Zheng, and Chen}]{DBLP:journals/corr/abs-2308-07269}
Peng Wang, Ningyu Zhang, Xin Xie, Yunzhi Yao, Bozhong Tian, Mengru Wang, Zekun Xi, Siyuan Cheng, Kangwei Liu, Guozhou Zheng, and Huajun Chen. 2023{\natexlab{a}}.
\newblock Easyedit: An easy-to-use knowledge editing framework for large language models.
\newblock \emph{CoRR}, abs/2308.07269.

\bibitem[{Wang et~al.(2023{\natexlab{b}})Wang, Li, Zhu, Zhang, Perera, Hang, Ma, Wang, Wang, Castelli, Xiang, and Ng}]{DBLP:conf/acl/WangLZZPHMWWCXN23}
Sijia Wang, Alexander~Hanbo Li, Henghui Zhu, Sheng Zhang, Pramuditha Perera, Chung{-}Wei Hang, Jie Ma, William~Yang Wang, Zhiguo Wang, Vittorio Castelli, Bing Xiang, and Patrick Ng. 2023{\natexlab{b}}.
\newblock Benchmarking diverse-modal entity linking with generative models.
\newblock In \emph{{ACL} (Findings)}, pages 7841--7857. Association for Computational Linguistics.

\bibitem[{Wang et~al.(2023{\natexlab{c}})Wang, Lv, Yu, Hong, Qi, Wang, Ji, Yang, Zhao, Song, Xu, Xu, Li, Dong, Ding, and Tang}]{DBLP:journals/corr/abs-2311-03079}
Weihan Wang, Qingsong Lv, Wenmeng Yu, Wenyi Hong, Ji~Qi, Yan Wang, Junhui Ji, Zhuoyi Yang, Lei Zhao, Xixuan Song, Jiazheng Xu, Bin Xu, Juanzi Li, Yuxiao Dong, Ming Ding, and Jie Tang. 2023{\natexlab{c}}.
\newblock Cogvlm: Visual expert for pretrained language models.
\newblock \emph{CoRR}, abs/2311.03079.

\bibitem[{Wei et~al.(2022)Wei, Tay, Bommasani, Raffel, Zoph, Borgeaud, Yogatama, Bosma, Zhou, Metzler, Chi, Hashimoto, Vinyals, Liang, Dean, and Fedus}]{DBLP:journals/tmlr/WeiTBRZBYBZMCHVLDF22}
Jason Wei, Yi~Tay, Rishi Bommasani, Colin Raffel, Barret Zoph, Sebastian Borgeaud, Dani Yogatama, Maarten Bosma, Denny Zhou, Donald Metzler, Ed~H. Chi, Tatsunori Hashimoto, Oriol Vinyals, Percy Liang, Jeff Dean, and William Fedus. 2022.
\newblock Emergent abilities of large language models.
\newblock \emph{Trans. Mach. Learn. Res.}, 2022.

\bibitem[{Wei et~al.(2023)Wei, Feng, Sun, Wang, Guo, and Yin}]{DBLP:conf/cvpr/WeiFS0GY23}
Qi~Wei, Lei Feng, Haoliang Sun, Ren Wang, Chenhui Guo, and Yilong Yin. 2023.
\newblock Fine-grained classification with noisy labels.
\newblock In \emph{{CVPR}}, pages 11651--11660. {IEEE}.

\bibitem[{Yao et~al.(2023)Yao, Wang, Tian, Cheng, Li, Deng, Chen, and Zhang}]{DBLP:conf/emnlp/YaoWT0LDC023}
Yunzhi Yao, Peng Wang, Bozhong Tian, Siyuan Cheng, Zhoubo Li, Shumin Deng, Huajun Chen, and Ningyu Zhang. 2023.
\newblock Editing large language models: Problems, methods, and opportunities.
\newblock In \emph{{EMNLP}}, pages 10222--10240. Association for Computational Linguistics.

\bibitem[{Zeng et~al.(2023)Zeng, Chen, Xu, and Kalantari}]{DBLP:conf/siggrapha/ZengCXK23}
Libing Zeng, Lele Chen, Yi~Xu, and Nima~Khademi Kalantari. 2023.
\newblock Mystyle++: {A} controllable personalized generative prior.
\newblock In \emph{{SIGGRAPH} Asia}, pages 70:1--70:11. {ACM}.

\bibitem[{Zhang et~al.(2024)Zhang, Yao, Tian, Wang, Deng, Wang, Xi, Mao, Zhang, Ni, Cheng, Xu, Xu, Gu, Jiang, Xie, Huang, Liang, Zhang, Zhu, Zhou, and Chen}]{DBLP:journals/corr/abs-2401-01286}
Ningyu Zhang, Yunzhi Yao, Bozhong Tian, Peng Wang, Shumin Deng, Mengru Wang, Zekun Xi, Shengyu Mao, Jintian Zhang, Yuansheng Ni, Siyuan Cheng, Ziwen Xu, Xin Xu, Jia{-}Chen Gu, Yong Jiang, Pengjun Xie, Fei Huang, Lei Liang, Zhiqiang Zhang, Xiaowei Zhu, Jun Zhou, and Huajun Chen. 2024.
\newblock A comprehensive study of knowledge editing for large language models.
\newblock \emph{CoRR}, abs/2401.01286.

\bibitem[{Zhang et~al.(2023)Zhang, Gu, and Plummer}]{DBLP:conf/emnlp/ZhangGP23}
Zhongping Zhang, Yiwen Gu, and Bryan~A. Plummer. 2023.
\newblock Show, write, and retrieve: Entity-aware article generation and retrieval.
\newblock In \emph{{EMNLP} (Findings)}, pages 8684--8704. Association for Computational Linguistics.

\bibitem[{Zheng et~al.(2023)Zheng, Li, Dong, Fan, Wu, Xu, and Chang}]{DBLP:conf/emnlp/ZhengLDFWXC23}
Ce~Zheng, Lei Li, Qingxiu Dong, Yuxuan Fan, Zhiyong Wu, Jingjing Xu, and Baobao Chang. 2023.
\newblock Can we edit factual knowledge by in-context learning?
\newblock In \emph{{EMNLP}}, pages 4862--4876. Association for Computational Linguistics.

\bibitem[{Zhou et~al.(2023)Zhou, Li, Xiang, Yan, Gui, and He}]{DBLP:journals/corr/abs-2311-00237}
Yuxiang Zhou, Jiazheng Li, Yanzheng Xiang, Hanqi Yan, Lin Gui, and Yulan He. 2023.
\newblock The mystery and fascination of llms: {A} comprehensive survey on the interpretation and analysis of emergent abilities.
\newblock \emph{CoRR}, abs/2311.00237.

\bibitem[{Zhu et~al.(2020)Zhu, Rawat, Zaheer, Bhojanapalli, Li, Yu, and Kumar}]{DBLP:journals/corr/abs-2012-00363}
Chen Zhu, Ankit~Singh Rawat, Manzil Zaheer, Srinadh Bhojanapalli, Daliang Li, Felix~X. Yu, and Sanjiv Kumar. 2020.
\newblock Modifying memories in transformer models.
\newblock \emph{CoRR}, abs/2012.00363.

\bibitem[{Zhu et~al.(2023)Zhu, Chen, Shen, Li, and Elhoseiny}]{DBLP:journals/corr/abs-2304-10592}
Deyao Zhu, Jun Chen, Xiaoqian Shen, Xiang Li, and Mohamed Elhoseiny. 2023.
\newblock Minigpt-4: Enhancing vision-language understanding with advanced large language models.
\newblock \emph{CoRR}, abs/2304.10592.

\bibitem[{Zou et~al.(2023)Zou, Chen, Shi, Guo, and Ye}]{DBLP:journals/pieee/ZouCSGY23}
Zhengxia Zou, Keyan Chen, Zhenwei Shi, Yuhong Guo, and Jieping Ye. 2023.
\newblock Object detection in 20 years: {A} survey.
\newblock \emph{Proc. {IEEE}}, 111(3):257--276.

\end{thebibliography}
\bibliographystyle{acl_natbib}

\end{document}